\documentclass[letterpaper, 10 pt, conference]{ieeeconf}  

\IEEEoverridecommandlockouts                              

\usepackage[table,xcdraw]{xcolor}
\usepackage{graphics} 
\usepackage{svg}
\usepackage{epsfig} 
\usepackage{mathptmx} 
\usepackage{times} 
\usepackage{amsmath} 
\usepackage{amssymb}  
\usepackage{cite}

\usepackage{hyperref}
\hypersetup{%
  breaklinks,
  bookmarksnumbered=true,
  bookmarksopen=true,
  colorlinks=true,
  linkcolor=black,
  urlcolor=black,
  citecolor=black
}

\usepackage{cleveref}
\usepackage{colortbl}
\usepackage{graphicx}
\usepackage{float}
\usepackage{comment}
\usepackage{booktabs}
\usepackage{subcaption}
\usepackage{siunitx}

\title{\textbf{
When to Waddle: A Comparative Study of Bipedal Torso-Stabilization on Low-Friction Surfaces
}}

\author{Ben Gu$^{1^*}$, Naomi Oke$^{2^*}$, George Ortiz$^{2}$, Stacy Ashlyn$^{3}$, Cordelia Pride$^{2}$,\\
Sarah Bergbreiter$^{2}$, Aaron M. Johnson$^{1,2}$ %
\thanks{*Equal contribution.}
\thanks{This work used Bridges-2 at Pittsburgh Supercomputing Center through allocation CIS240861 from the Advanced Cyberinfrastructure Coordination Ecosystem: Services \& Support (ACCESS) program, which is supported by U.S. National Science Foundation grants \#2138259, \#2138286, \#2138307, \#2137603, and \#2138296. This work was supported in part by the National Science Foundation under Grant CMMI-2408884. Any opinions, findings, and conclusions or recommendations expressed in this material are those of the author(s) and do not necessarily reflect the views of the National Science Foundation.}
\thanks{$^{1}$Robotics Institute, Carnegie Mellon University, Pittsburgh, PA 15213, USA {\tt\small bengu@andrew.cmu.edu}}%
\thanks{$^{2}$Department of Mechanical Engineering, Carnegie Mellon University, Pittsburgh, PA 15213, USA {\tt\small noke@andrew.cmu.edu}}%
\thanks{$^{3}$Mechanical and Aerospace Engineering, New York University, New York, NY 10012, USA} 
}

\begin{document}

\maketitle
\begin{abstract}
Low-friction surfaces challenge bipedal locomotion by limiting the contact forces available during stepping. Inspired by penguin waddling, we investigate how lateral torso motion and center of mass (COM) placement affect locomotion as surface friction changes. Using a five-actuator biped, we compare an upright-gait strategy with a penguin-inspired torso-over-stance-leg strategy across multiple COM placements in simulation and hardware. In the 3-D simulator MuJoCo, we sweep through sinusoidal leg and hip actuation parameters across four friction coefficients $\mu \in \{0.1, 0.3, 0.5, 0.7\}$. In simulation, torso-over-stance-leg motion produces more successful controllers and higher forward speeds at low friction, with the highest speed occurring for the high-COM configuration. Hardware experiments show the same low-friction speed trend: at $\mu=0.12$, torso-over-stance-leg motion increases forward speed and reduces cost of transport at both tested COM ratios, and the higher COM also improves both measures. The high-COM penguin configuration is the fastest and most energy efficient while maintaining low sideways foot motion. At $\mu=0.45$, the COM trend reverses: the lower-COM configurations are faster and more energy efficient, while gait strategy has little effect on forward speed but still changes sideways foot motion. These results show that the effects of lateral torso motion and COM placement depend on the available friction, and that forward speed, energy use, and slip-related foot motion can be modulated with a penguin-inspired torso motion on hardware.
\end{abstract}

\section{Introduction}

Legged robots operating in real-world environments must maintain stable and efficient locomotion despite changes in terrain and contact conditions. Low-friction surfaces are particularly challenging for bipeds because stance-foot slip violates the no-slip contact assumption underlying many walking models and can rapidly destabilize locomotion. Previous work has addressed this problem through friction-aware gait and footstep planning \cite{brandao2014,brandao2016footstep}, vision-based surface recognition coupled with hierarchical planning \cite{brandao2016material}, and trajectory optimization that explicitly allows and plans stance-foot slip \cite{ma2019}. Other studies have investigated gait parameters that improve robustness to slipping \cite{chen2021slip} and developed whole-body or adaptive controllers that maintain balance after a slip occurs \cite{mihalec2022,mihalec2023,almeida2024}. More recently, learning-based humanoid controllers have demonstrated locomotion across slippery and other challenging terrains \cite{jung2025}. These approaches demonstrate that low-friction locomotion can be improved through perception, planning, and feedback control. While these approaches are reliable, we explore whether morphology and whole-body gait dynamics can themselves provide robustness as the available friction decreases.

Biological bipeds provide evidence that stability on uncertain surfaces emerges from the interaction between control and body mechanics. Guinea fowl traversing low-friction terrain adjust limb posture and the position of the center of mass (COM) relative to the supporting limb to limit slip and avoid falls \cite{clark2011}. Pacific parrotlets similarly maintain wing and leg dynamics when landing on surfaces with different geometries and textures, while adapting toe and claw interactions after contact \cite{roderick2019}. Avian-inspired robots have demonstrated that such mechanical specialization can reduce the control required for locomotion: BirdBot, for example, exploits avian-inspired leg morphology and passive mechanical coupling to produce energy-efficient walking with minimal control \cite{badri2022birdbot}.

\begin{figure}[t]
    \centering
    \includegraphics[width=0.98\columnwidth]{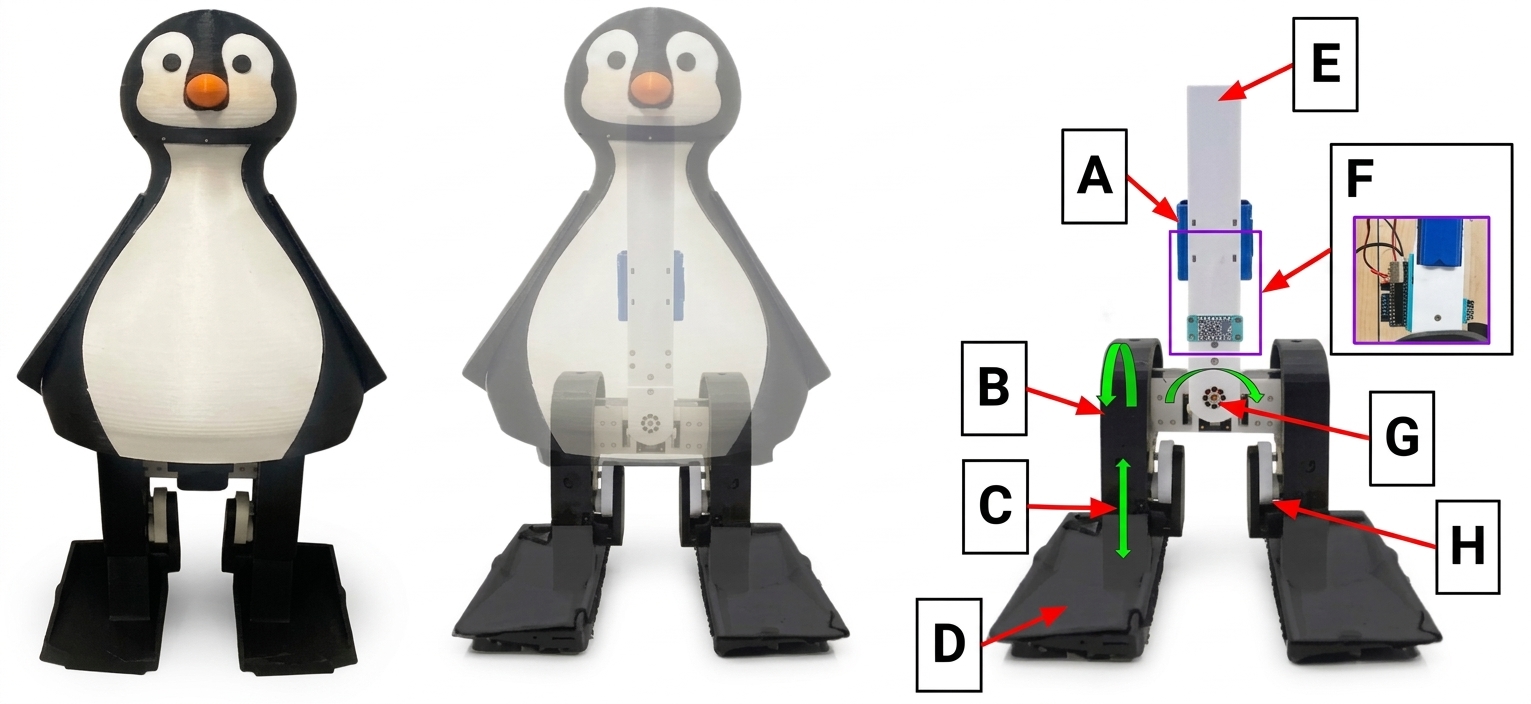}
    \caption{The Pengu platform: (A) 7.4\,V LiPo battery, (B) hip-pitch actuator, (C) leg-extension actuator, (D) ellipsoidal curved foot, (E) torso holding mass, (F) microcontroller (rear-mounted, shown inset) with BNO055 IMU (front-mounted), (G) torso-roll actuator, (H) crank-slider mechanism. Green arrows show the direction of the actuated DOF.}
    \label{fig:pengu_robot}
\end{figure}

Penguins provide a particularly interesting example of morphology-coupled bipedal locomotion. Their short limbs, comparatively large torso, and characteristic lateral body oscillation produce terrestrial dynamics different from conventional human walking on low-friction surfaces. In Emperor penguins, lateral waddling facilitates mechanical-energy exchange rather than simply adding energetic cost \cite{griffinkram2000}. In King penguins, step width is less variable than step length, suggesting that waddling can provide a repeatable strategy for regulating frontal-plane stability \cite{kurz2008}. Recent anatomical work on Macaroni penguins further identifies specialized hindlimb morphology, including a distinct \textit{m. adductor tibialis} that maintains the tibiotarsi in adduction and supports an otherwise abducted femoral posture during terrestrial stance \cite{hirsh2026}. This anatomical specialization is consistent with recent work investigating the penguin waddle as a morphology-driven gait and its translation to robot design \cite{ashlyn2024,ashlyn2026,oke2026}. Together, these studies motivate treating lateral body oscillation as a gait feature whose utility may depend on the contact environment.

Simple bipedal robots provide a useful platform for isolating such embodied effects. Quasi-passive walkers have demonstrated that carefully selected foot geometry, mass distribution, and actuation can produce stable three-dimensional locomotion with little or no feedback
\cite{mcgeer1990passive,collins2005efficient,islam2022,kyle2023}. Other bird-inspired robots like Microduck allow further insights into bipedal walking \cite{pollen2026microduck}. Terrestrial penguin-inspired robots, however, are not as common in literature. Toya and Ishikawa generated stable stepping in a penguin-inspired waddling mechanism using body sway and attitude stabilization \cite{toya2024}, and subsequently combined this stabilization strategy with reinforcement learning to generate forward locomotion in simulation \cite{toya2025}. Liu et al.\ developed a miniature underactuated tripod robot driven by an oscillatory mechanism inspired by penguin waddling \cite{liu2025}. These studies establish that penguin-like body oscillation can generate locomotion, but do not systematically isolate how penguin-inspired body dynamics compare with more human-like walking strategies as surface friction changes on hardware.

In this work, we present a self-stabilizing penguin-inspired bipedal robot that is \SI{54}{\centi\meter} tall with a \SI{15}{\centi\meter} hip height and investigate its locomotion across surfaces with different friction properties. We compare two torso-control gait strategies across different COM placements to determine how their performance changes with available friction: zero-roll torso tracking and penguin-inspired torso-over-stance-leg tracking. We characterize robustness, speed, energetic efficiency, body motion, and support behavior of walking to identify the dynamical mechanisms associated with successful low-friction locomotion. Beyond serving as a platform for studying embodied locomotion, robust small-scale walking robots could eventually support environmental monitoring in polar environments, where robotic observation has been shown to reduce disturbance to penguin colonies \cite{lemaho2014} and where climate pressures and emerging disease threats motivate improved remote monitoring \cite{gimeno2024,wille2025}. By connecting penguin biomechanics, existing biped design, and low-friction locomotion, this work investigates when and why it is advantageous for a robot to walk like a penguin.

\section{Bipedal Robot Design}
\subsection{Mechanical}

\begin{figure}[t]
  \centering
  \includegraphics[width=0.485\textwidth]{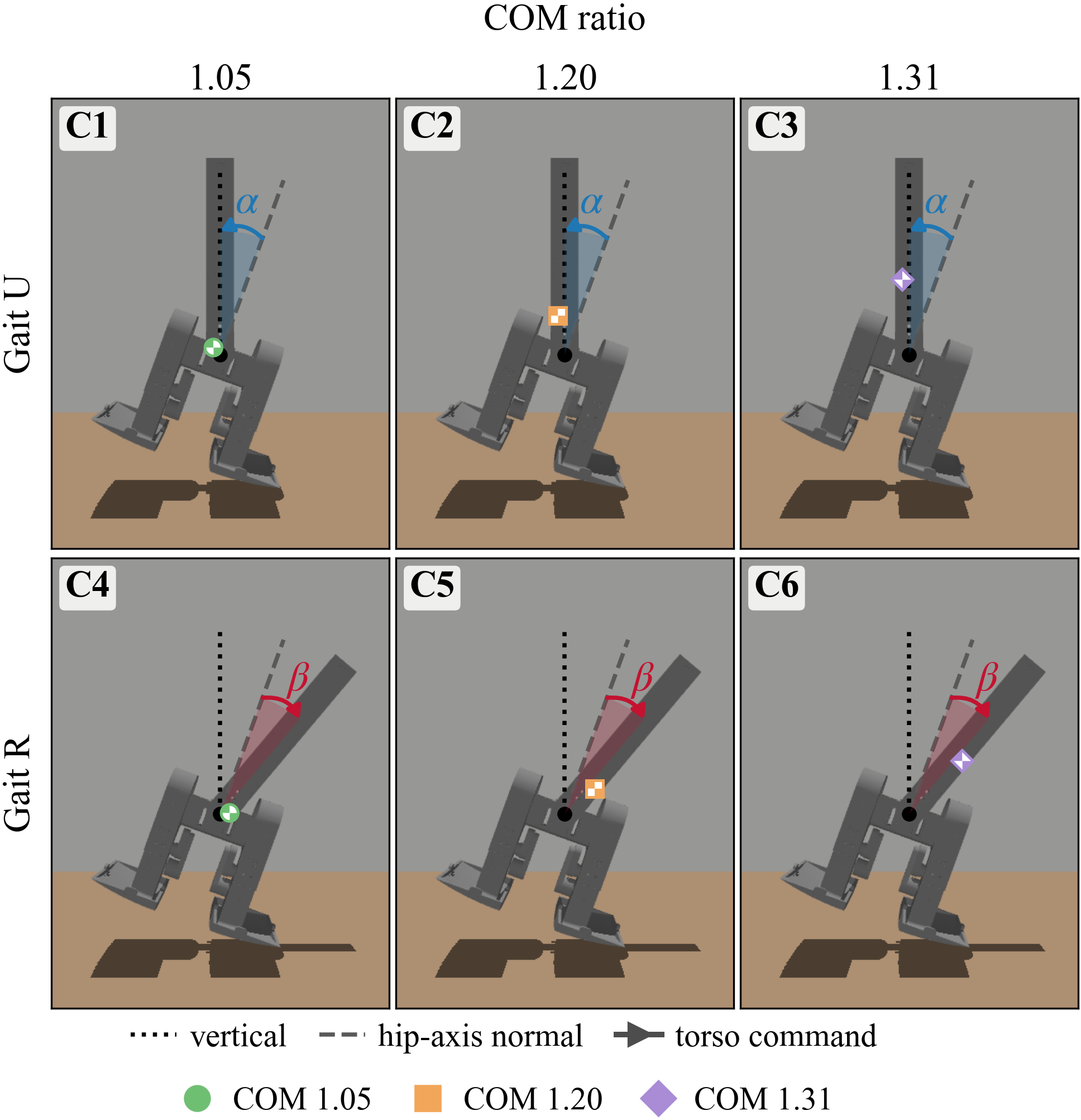}
  \caption{The six configurations: two gait strategies (Gait U, Gait R) $\times$ three COM ratios (markers, height exaggerated); the same robot model is drawn in every panel. $\alpha$ and $\beta$ depict the angle made between the current torso position and target position.}
  \label{fig:configs}
\end{figure}

The robot consists of six rigid bodies and five joints: a torso, hip axis, two upper thighs, and two telescoping lower legs with curved feet (\Cref{fig:pengu_robot}). The five degrees of freedom (DOFs) were selected as the minimum required to reproduce the Macaroni penguin's walking gait \cite{ashlyn2026}. All structural components were 3-D printed in PLA. To allow modeling a penguin-like mass distribution (70--80\% of the robot’s weight), the torso is hollow so weights can be added \cite{ashlyn2026}. This was done to capture the change in normal force caused by body motion. The COM is changed by placing titanium and steel weights at different heights in the torso, resulting in a total mass of \SI{2.29}{\kilogram}. 

Additionally, we created a 3-D printed penguin-inspired shell out of PLA for visualization. The shell was not used during experiments or included in the dynamics and gait analysis (\Cref{fig:pengu_robot}).

\subsection{Electronics}
The robot is actuated by Dynamixel XM430-W350-T motors. Two motors drive a crank-slider mechanism for prismatic leg extension, enabling approximately 4 cm of foot clearance. Two additional motors control hip motion, with another for torso roll. The torso DOF enables lateral weight shifting, which plays a key role in frontal and lateral stability. Passive lateral stability is achieved by ellipsoidal feet with non-concentric curvature centers \cite{islam2022}. The electronics housed on the torso for power and communication include an Arduino MKR WiFi 1010 with a Dynamixel Shield, dual 2S LiPo batteries, and an Adafruit BNO055 IMU.
Locomotion is generated using open-loop sinusoidal position control applied to all five actuators. The hip and leg motions share a common frequency, producing a symmetric gait with phase offsets $\theta_i=0^\circ$ and $180^\circ$ for the left and right legs, respectively,
\begin{align}
    p_i = A_i \sin(\omega t + \phi_i) + \theta_i 
    \label{eq:controller}
\end{align}
with separate amplitudes $A_i$ and offsets $\theta_i$ for each joint.



We explore the effects of mass distribution and gait strategy. We characterize the mass distribution by the \emph{COM ratio}, defined as whole-robot COM height divided by hip-axis height in the neutral standing pose. Three COM ratios are evaluated: a human-like COM, 1.05~\cite{deleva1996}; a Macaroni-penguin-like COM, 1.31~\cite{ashlyn2026}; and a COM ratio in-between both values, 1.20. The total mass (\SI{2.29}{\kilogram}), geometry, and actuation DOFs are otherwise held constant. We assess the gait strategy using two gaits: Gait U (upright) and Gait R (rolling). Gait U tracks zero world-frame torso roll to maintain an upright torso, while Gait R tracks the torso toward and over the stance leg. Unlike Gait U, Gait R uses the common frequency $\omega$ used by the hip and legs (\Cref{eq:controller}). Together, gait strategy and COM ratio define the six configurations in \Cref{fig:configs}.

\subsection{Simulation Methods}
We simulate the robot in MuJoCo (v3.8) from a CAD-derived model of the platform in \Cref{fig:pengu_robot}. The model preserves the closed crank--shaft leg mechanism: each leg's telescopic extension is a passive prismatic joint closed through a crank linkage by equality constraints, so the five position-servo actuators in simulation match the five physical motors (two hip, one crank disk per leg, and one torso-roll actuator). Actuator torque is clamped to the Dynamixel XM430-W350-T stall torque ($\pm\SI{4.1}{\newton\meter}$), and the simulation uses a \SI{1}{\milli\second} timestep with MuJoCo's \texttt{implicitfast} integrator. The ground friction coefficient $\mu$ is imposed through the foot--floor contact parameters.

The legs and hips follow open-loop sinusoidal control \eqref{eq:controller}, modeling the actuation parameters of the hardware. The torso DOF motion is closed-loop based on the feedback from a simulated IMU.


\subsection{Gait--Friction Sweep}
For each configuration, we sweep the five open-loop gait parameters defined in \Cref{tab:sweep}. This is done across four friction coefficients, $\mu \in \{0.1, 0.3, 0.5, 0.7\}$. Each trial is a deterministic \SI{13}{\second} rollout after \SI{2}{\second} of settle time.

We define a controller as \emph{successful} if the robot does not fall, travels faster than \SI{0.05}{\meter\per\second}, keeps its body heading within \ang{90} of its direction of travel, keeps the commanded servo velocities below the \ang{354}/s limit for at least three quarters of the gait cycle. Additionally, we filter the controllers to walking gaits by excluding rollouts with a flight phase.

A controller is defined as \emph{robust} if all of its immediate neighbors are successful i.e., one grid step in stride frequency ($\pm\SI{0.05}{\hertz}$) and one in hip phase ($\pm\ang{10}$). The robust region measures the number of robust controllers out of the full grid sweep.

For each configuration and friction coefficient, we compare the size of the robust region, the forward speed, the body--contact geometry, and the support-phase behavior. Because flight phases are excluded, the resulting gaits are alternating and the duty factor $D$ converts directly to a
double-support fraction $2D-1$ and a single-support fraction $2(1-D)$. The highest-performing controllers are then compared across the six configurations at each friction coefficient.

\begin{table}[t]
\centering
\caption{Controller sweep grid}
\label{tab:sweep}
\begin{tabular}{lrr}
\toprule
Parameter & Range & Step \\
\midrule
Stride frequency $f$ [\si{\hertz}] & 1.20--2.00 & 0.05 \\
Hip phase $\phi$ [\si{\degree}]    & 0--350     & 10   \\
Leg amplitude $A_{leg}$ [\si{\degree}]                  & 70--130    & 5    \\
Hip amplitude $A_{hip}$ [\si{\degree}]                      & 12--32     & 4    \\
Hip offset                         & 20--40     & 5    \\
\bottomrule
\end{tabular}
\end{table}
\begin{figure*}[t]
    \centering
    \includegraphics[width=0.485\textwidth]{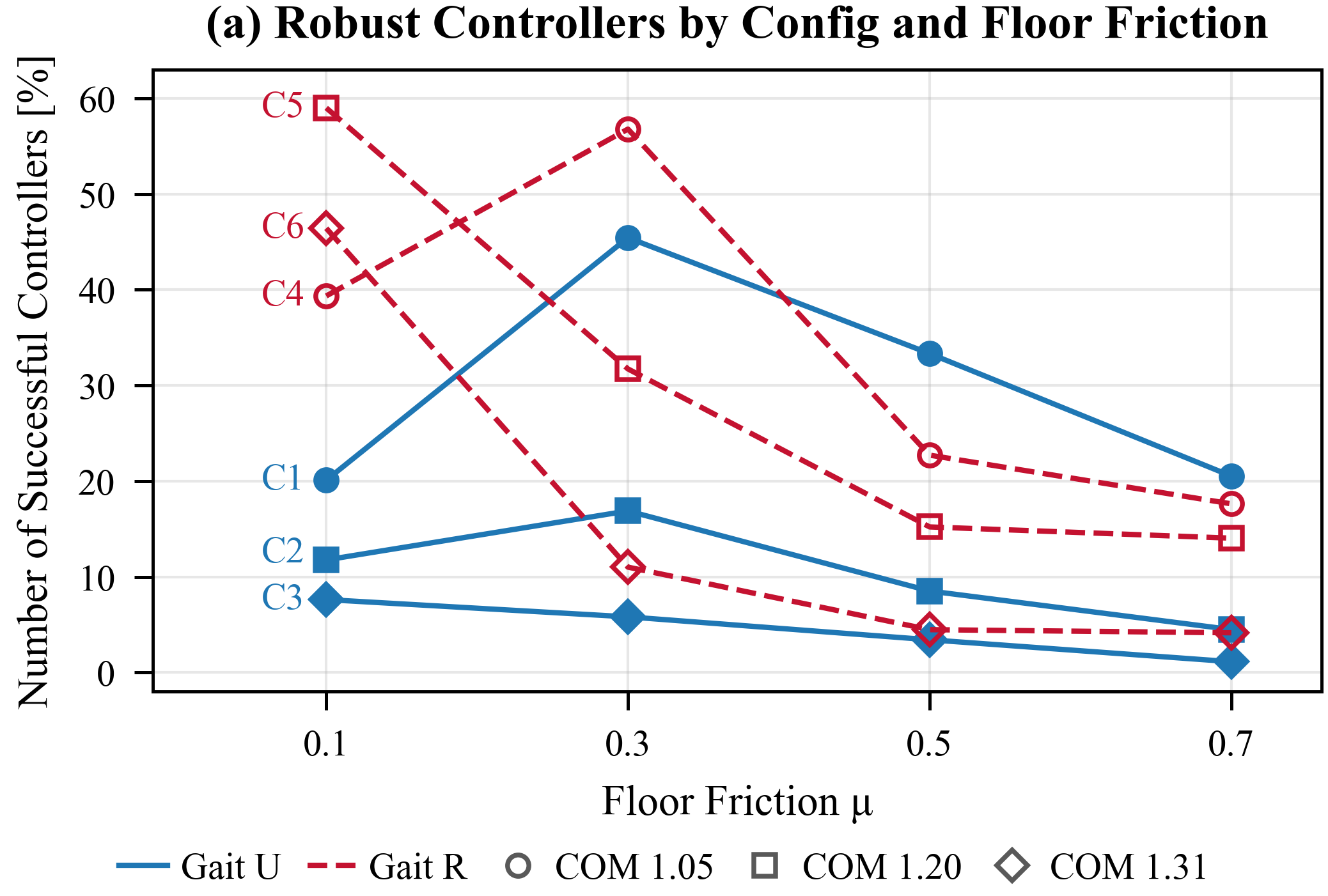}%
    \hfill
    \includegraphics[width=0.485\textwidth]{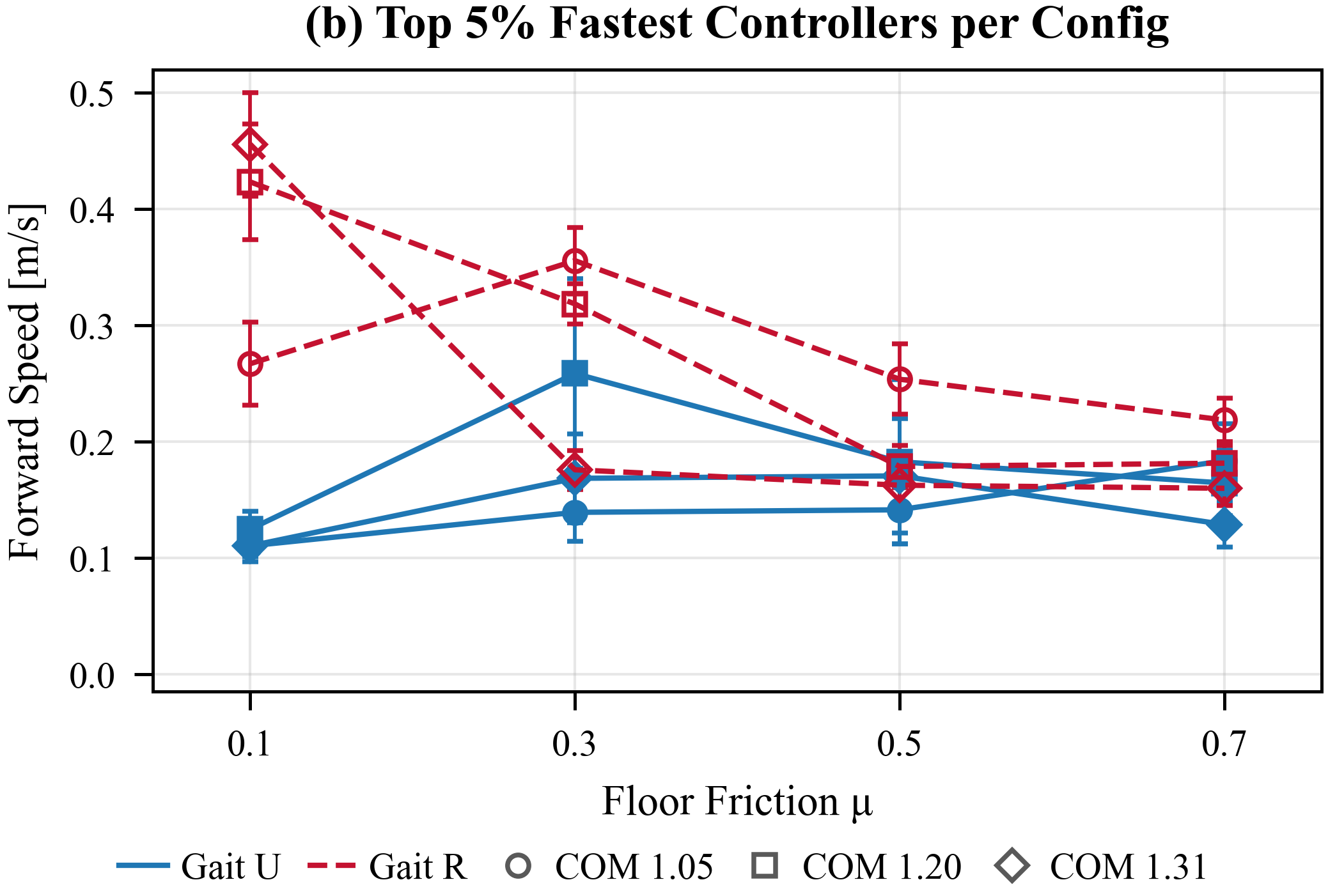}
    \caption{Simulation performance across friction coefficients. (a) Number of robust controllers for each configuration. (b) Forward speed of the top 5\% of successful controllers.}
    \label{fig:sim_performance}
\end{figure*}

\subsection{Hardware Experimental Methods}


The physical robot was evaluated on a rubber floor with a coefficient of friction $\mu=0.45$ and a recycled slippery UHMW polyethylene sheet $\mu=0.12$. Hardware experiments crossed two COM ratios with the two gait strategies: C1 (COM ratio = 1.05, Gait U), C3 (COM ratio = 1.31, Gait U), C4 (COM ratio = 1.05, Gait R), and C6 (COM ratio = 1.31, Gait R). Body motion was recorded using motion capture, and electrical power was recorded using an external power supply.

Actuation parameters from the simulated robust region were used for hardware experiments. The actuator parameters were manually tuned until stable walking was achieved. The simulated and hand-tuned values are described in \Cref{tab:locomotion_metrics}.


Whole-body COM trajectories were reconstructed from motion-capture marker trajectories. Segment COM locations were obtained from the CAD-derived mass model and transformed into the motion-capture frame from each segment's marker cluster. Whole-body COM was calculated as
\begin{equation}
    \mathbf{x}_{\mathrm{COM}}(t)
    =
    \frac{\sum_i m_i \mathbf{x}_i(t)}
         {\sum_i m_i}.
\end{equation}

The robot's forward direction was obtained from the torso orientation and projected in the horizontal plane. The sideways direction was defined perpendicular to this heading. 

The reported forward speed was calculated as
\begin{equation}
    v_{\mathrm{fwd}}
    =
    \frac{\sum_k \Delta x_{\mathrm{fwd},k}}
         {\sum_k \Delta t_k},
\end{equation}
where $\Delta x_{\mathrm{fwd},k}$ is the forward COM displacement over stride $k$. Sideways COM speed was computed from the magnitude of the displacement perpendicular to the torso heading over each stride and averaged across strides.

Electrical power was taken from the power-supply recording. Cost of transport was calculated as
\begin{equation}
    \mathrm{COT}
    =
    \frac{P}{mgv_{\mathrm{fwd}}},
\end{equation}
where $P$ is electrical power, $m$ is robot mass, and $g=\SI{9.81}{\meter\per\second\squared}$. Values are reported as mean $\pm$ standard deviation across retained trials.


\section{Results}

\subsection{Simulation}

First, we test how gait strategy affects both the robustness (i.e.\ the size of the successful gait region) and the achieved forward speed.  At \(\mu=0.1\), Gait R (torso-over-stance-leg) produced more successful controllers than Gait U (upright torso) at all three COM ratios. The COM trend depended on gait strategy at this friction: Gait U favored the lower COM (C1), while Gait R produced the most successful controllers at the intermediate COM ratio (C5), followed by the higher COM ratio (C6). As friction increased, lower COM ratios were generally associated with more successful controllers for both gaits, and the difference between gaits was less pronounced. Forward speed shows a related low-friction trend. At \(\mu=0.1\), Gait R is faster than Gait U at each COM ratio, and the high-COM Gait R configuration reaches the highest mean speed (\Cref{fig:sim_performance}). The Gait R speed advantage becomes smaller and more dependent on COM ratio as friction increases.

\begin{figure*}[h]
    \centering
    \includegraphics[width=0.98\textwidth]{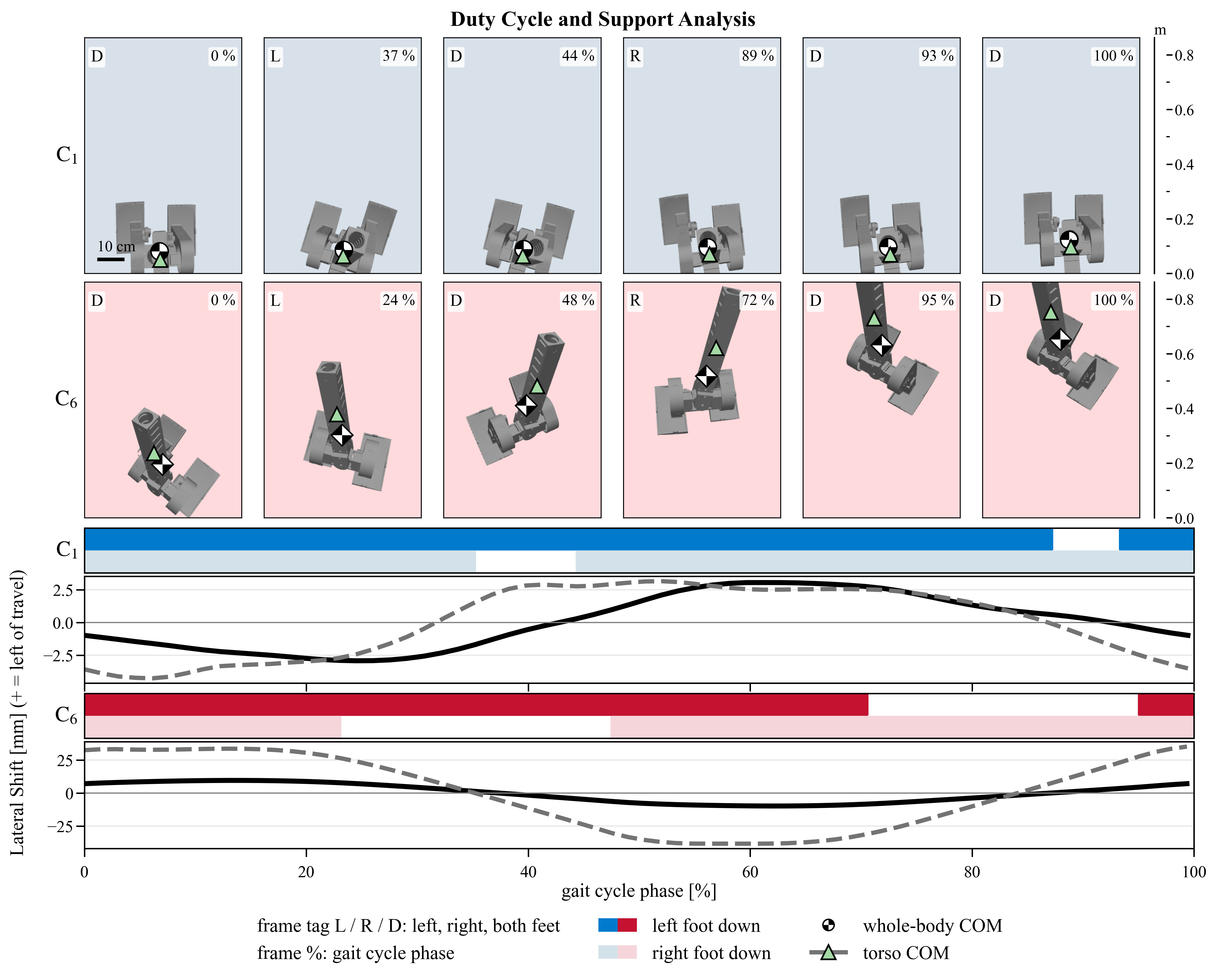}
    \caption{Duty cycle and support analysis for C1 (Gait U) and C6 (Gait R) at $\mu = 0.1$.
    Top: one gait cycle for each configuration, seen from above, at the cycle start, the four contact events (right touchdown, left toe-off, left touchdown, right toe-off) and the cycle end; the ruler indicates the vertical distance spanned by each frame at floor level. Bottom: which foot is in contact over the cycle, and the lateral shift of the whole-body and torso COM.}
    \label{fig:sim_dutyfactor}
\end{figure*}

At $\mu=0.1$, C1 spends most of the gait cycle in double support, with only short intervals of single support (\Cref{fig:sim_dutyfactor}). C6 shows a more alternating support pattern, with longer single-support periods for each leg and shorter double-support transitions. Additionally, C6 has larger lateral torso motion.

\newcommand{\goodcell}{\cellcolor{green!15}}
\newcommand{\midcell}{}
\newcommand{\badcell}{\cellcolor{red!15}}

\subsection{Hardware Experimentation}

Hardware experiments crossed two COM ratios with the two gait strategies: C1 and C3 use Gait U with COM ratios 1.05 and 1.31, respectively, while C4 and C6 use Gait R with the same COM ratios (\Cref{fig:configs}). Results for the two friction conditions are summarized in \Cref{tab:locomotion_metrics}.

\subsubsection{Low-Friction Surface}

Raising the COM improves speed and COT for both gait strategies, and switching from Gait U to Gait R improves speed and COT for both COM ratios.

For Gait U, increasing the COM ratio from 1.05 to 1.31 increased forward speed from $0.145$ to $0.227$ \si{\meter\per\second} and reduced COT from $6.4$ to $4.6$. For Gait R, the same change increased forward speed from $0.199$ to $0.261$ \si{\meter\per\second} and reduced COT from $5.1$ to $3.6$. 

Gait R also improved forward speed and COT at both COM ratios compared to Gait U. At the lower COM ratio, C4 was faster and had lower COT than C1. At the higher COM ratio, C6 was faster and had lower COT than C3. C6 was the fastest and most energy-efficient configuration measured on the low-friction surface.

Sideways foot speed depended on both COM placement and gait strategy. C3 had the largest sideways foot speed, $0.088\pm0.017$ \si{\meter\per\second}, while C6 had the smallest, $0.032\pm0.015$ \si{\meter\per\second}, even though both used the higher COM ratio. 

At $\mu=0.12$, the higher COM improved forward speed and energetic cost for both gait strategies (\Cref{tab:locomotion_metrics}). At the lower COM ratio, gait strategy had little effect on sideways foot speed. Lower COM ratios C1 and C4 had similar sideways foot speeds of $0.037\pm0.010$ and $0.041\pm0.010$ \si{\meter\per\second}, respectively.

\subsubsection{Higher-Friction Surface}

At $\mu=0.45$, the effect of COM placement reversed (\Cref{tab:locomotion_metrics}). The low-COM configurations C1 and C4 reached nearly identical forward speeds of $0.107$ and $0.108$ \si{\meter\per\second}, while the high-COM configurations C3 and C6 reached only $0.044$ and $0.043$ \si{\meter\per\second}. COT showed the same separation: C1 and C4 had COT values of $8.9$ and $7.8$, compared with $23.6$ and $24.6$ for C3 and C6.

Gait strategy had little effect on forward speed within a given COM ratio at higher friction. However, Gait R reduced sideways foot speed at both COM ratios. Sideways foot speed decreased from $0.035$ to $0.024$ \si{\meter\per\second} between C1 and C4, and from $0.018$ to $0.009$ \si{\meter\per\second} between C3 and C6. The high-COM configurations therefore had the lowest sideways foot motion on this surface while also having the lowest forward speeds and highest COT.



\subsection{Simulation-to-Hardware Comparison}

At low friction, simulation and hardware agree on the direction of the gait strategy effect. In simulation at $\mu=0.1$, Gait R is faster than Gait U for each COM ratio, and the high-COM Gait R configuration reaches the highest forward speed. Hardware results at $\mu=0.12$ follow the same trend.

The simulation also produces more successful Gait R controllers than Gait U controllers at low friction. Because success is evaluated over a controller sweep with friction and initial-pose perturbations, the larger successful region suggests that Gait R may tolerate a wider range of low-friction conditions in simulation. The hardware experiments were not designed to estimate a success rate over comparable perturbations, allowing for future robustness experimentation. 

At higher friction, the hardware performance changes more by COM placement. C1 and C4 have nearly identical forward speeds, as do C3 and C6, while changing the COM ratio produces a large change in speed and COT.  The simulation also shifts away from the high-COM low-friction optimum as friction increases. Near \(\mu=0.5\), the simulated gait strategy effect is COM-dependent. Gait R retains a speed advantage at the lower COM, while the two strategies are similar or reverse ordering at the higher COM ratios. On hardware at \(\mu=0.45\), gait strategy has little effect on forward speed at either tested COM ratio.
Differences in foot contact, actuator behavior, and unmodeled compliance may contribute to this gap, which can be explored further in future hardware experimentation. 

\begin{table*}[t]
    \centering
    \caption{Locomotion metrics on hardware at $\mu=0.12$ and $\mu=0.45$ (mean $\pm$ SD across 10--20 trials per configuration). Within each friction level, green marks the best hardware value (highest forward speed; lowest foot sideways speed; lowest COT) and red the worst; where the runner-up overlaps the extreme within one SD, both are shaded to indicate.}
    \label{tab:locomotion_metrics}
    \small
    \setlength{\tabcolsep}{5pt}
    \begin{tabular}{lccccc}
        \toprule
        Config ($\mu$) & Gait strategy & COM ratio & Fwd.\ speed & Foot sideways & COT \\
        & & & (\si{\meter\per\second}) & (\si{\meter\per\second}) & ($-$) \\
        \midrule

        \multicolumn{6}{l}{$0.12$} \\
        \midrule
        C1 & Gait U & 1.05
        & \badcell $0.145\pm0.007$
        & \goodcell $0.037\pm0.010$
        & \badcell $6.41\pm0.70$ \\

        C3 & Gait U & 1.31
        & \goodcell $0.227\pm0.038$
        & \badcell $0.088\pm0.017$
        & \goodcell $4.55\pm1.05$ \\

        C4 & Gait R & 1.05
        & \midcell $0.199\pm0.028$
        & \midcell $0.041\pm0.010$
        & \midcell $5.14\pm0.70$ \\

        C6 & Gait R & 1.31
        & \goodcell $0.261\pm0.019$
        & \goodcell $0.032\pm0.015$
        & \goodcell $3.56\pm0.45$ \\

        \midrule
        \multicolumn{6}{l}{$0.45$} \\
        \midrule
        C1 & Gait U & 1.05
        & \goodcell $0.107\pm0.011$
        & \badcell $0.035\pm0.016$
        & \goodcell $8.95\pm0.53$ \\

        C3 & Gait U & 1.31
        & \badcell $0.044\pm0.003$
        & \midcell $0.018\pm0.003$
        & \badcell $23.58\pm1.05$ \\

        C4 & Gait R & 1.05
        & \goodcell $0.108\pm0.006$
        & \badcell $0.024\pm0.004$
        & \goodcell $7.85\pm1.42$ \\

        C6 & Gait R & 1.31
        & \badcell $0.043\pm0.002$
        & \goodcell $0.009\pm0.002$
        & \badcell $24.58\pm1.29$ \\

        \bottomrule
    \end{tabular}
\end{table*}

One possible explanation is that some lateral foot motion on the low-friction surface allows the feet to accommodate the imposed open-loop motion and curved-foot contact. On the higher-friction surface, the same motion may generate larger lateral constraints or losses. The high-COM configurations show their sideways foot speed dropping strongly as friction increases, while their forward speed also drops and COT rises. C6 also shows that large sideways foot motion is not required for fast low-friction walking, since it combines low sideways foot speed with the best forward and energetic performance.

\subsection{When Should a Robot Walk Like a Penguin?}

For the hardware conditions tested here, the clearest benefit of Gait R appears at low friction. At $\mu=0.12$, it increases forward speed and reduces COT at both COM ratios, and the high-COM Gait R configuration performs best in all three reported metrics. At $\mu=0.45$, Gait R continues to reduce sideways foot speed, while forward speed is set mainly by COM placement and the COT change is small compared with the effect of COM ratio.

The simulation supports the low-friction speed benefit and also shows a larger successful controller region for Gait R. This gives a simulation-based reason to study penguin-like torso motion as a low-friction strategy, while the hardware results show that its measured benefit depends on both friction and mass distribution.

\section{Conclusion}

Penguin-inspired torso-over-stance-leg motion improved the low-friction hardware gait for both COM placements. At $\mu=0.12$, Gait R increased forward speed and reduced COT at both COM ratios, and C6 combined the highest forward speed, lowest COT, and lowest sideways foot speed. The higher COM also improved speed and COT on this surface. At $\mu=0.45$, the COM trend reversed and the lower-COM configurations were faster and more energy efficient, while Gait R mainly reduced sideways foot motion with little change in forward speed.

The simulation shows the same low-friction speed advantage for Gait R and a larger region of successful Gait R controllers. The success-rate result suggests greater low-friction tolerance in simulation, but hardware robustness was not tested. Across the hardware experiments, forward speed, energetic cost, and sideways foot motion respond differently to friction and COM placement. Future experiments should directly measure ground-reaction forces and stance-foot slip and test intermediate COM placements to determine how weight transfer, contact forces, and foot mechanics produce these changes.






\section*{ACKNOWLEDGMENT}
We would like to thank Hopewell Feldmann, Steven Man, and Nicole Ortuna for their contributions to this project. 
\textit{Disclosure per IEEE-RAS AI policy}: For this paper, Claude and ChatGPT were used to debug simulation code and to generate and debug figure-plotting code.


\bibliographystyle{IEEEtran}
\bibliography{references}

\end{document}